\documentclass[conference]{IEEEtran}
\IEEEoverridecommandlockouts

\usepackage{amsmath,amssymb,amsthm}
\usepackage{graphicx}
\usepackage{booktabs}
\usepackage[hidelinks]{hyperref}
\usepackage{xcolor}
\definecolor{cbblue}{RGB}{59,130,246}
\definecolor{cbbluedk}{RGB}{37,99,235}
\definecolor{cbgold}{RGB}{234,179,8}
\definecolor{cbgolddk}{RGB}{180,130,0}
\definecolor{cbred}{RGB}{220,38,38}
\definecolor{cbgreen}{RGB}{22,163,74}
\usepackage{enumitem}
\usepackage{microtype}
\usepackage{cite}
\usepackage{balance}
\usepackage{fancyhdr}
\usepackage{pgfplots}
\pgfplotsset{compat=1.18}
\usepackage{multirow}
\usepackage{caption}
\usepackage{subcaption}
\usepackage{array}
\usepackage{tabularx}
\usepackage{float}
\usepackage{placeins}

\title{Dynamic Governance of Multi-LLM Agent Systems\\
for Collaborative Conversational Outcomes}

\author{
  \IEEEauthorblockN{Alexander Liss}
  \IEEEauthorblockA{Georgia Institute of Technology\\\texttt{aliss6@gatech.edu}}
  \and
  \IEEEauthorblockN{Nicholas Desmond}
  \IEEEauthorblockA{Huge Inc.\\\texttt{ndesmond@hugeinc.com}}
  \and
  \IEEEauthorblockN{Santiago Gil Gallego}
  \IEEEauthorblockA{Huge Inc.\\\texttt{sgallego@hugeinc.com}}
}

\begin{document}

\maketitle
\setcounter{page}{1}
\pagestyle{fancy}
\fancyhf{}
\fancyfoot[C]{\thepage}
\renewcommand{\headrulewidth}{0pt}

\begin{abstract}
Classical multi-agent reinforcement learning composes a shared policy through
joint reward optimization. LLM agents lack this foundation: deployed in
multi-agent settings with structurally opposed objectives, they drift toward
attractor states rather than converging to cooperative equilibria. This paper
asks whether a control theory-informed governance layer can substitute for the missing
goal function, steering two LLM agents toward a jointly optimal outcome.

We propose a framework to address these challenges. The \textit{Experience Orchestrator} (EO) explores a simulated
financial services environment where a site agent guides a visitor toward
a "speak with financial advisor" digital action while the visitor agent maintains realistic
resistance given its persona. EO governs the joint trajectory through Contextual Bandit (CB)
action selection calibrated from real-world web session analytics, 
PID-controlled schema constraints, and POMDP belief tracking.

Across a 60,000-simulation factorial evaluation, the full system achieves a
\textbf{+32 point lift} in high-intent advisor contact rate
(78.1\% vs.\ 46.1\%) over an LLM control guided purely with a system prompt. Critically, CB variant selection accounts for
\textbf{97\% of between-factor outcome variance} versus 3\% for friction model
choice, confirming that the governance policy, not environmental initial
conditions, determines where trajectories end up.

These results carry important limitations: all findings derive from LLM-to-LLM
simulation rather than live human interaction, and the PID controller has not
been calibrated against the far greater unpredictability of real human visitors.
Validating EO in a live production environment and extending the governance
framework to the broader challenge of calibrating independent LLM agents toward
shared goals are the critical next steps for this research program.
\end{abstract}

\section{Introduction and Hypothesis}
\label{sec:intro}

\subsection{The Missing Goal Function}

The challenge of coordinating independent LLM agents toward a shared objective
has emerged as one of the most pressing problems in applied AI. Multi-agent
systems are being deployed at scale across business domains, from customer
service and sales automation to enterprise workflow orchestration and clinical
decision support. Despite this rapid adoption, the fundamental question of how
to make structurally independent agents collaborate effectively remains largely
unsolved. There has been extensive prior work on multi-agent reinforcement
learning in classical settings, but the shift to LLM agents breaks the
assumptions that prior work relied on.

Classical RL agents optimize an explicit
reward function; the goal is mathematically encoded in every gradient update.
LLM agents have no equivalent. Their behavior is shaped by a prompt, which
specifies intent in natural language but provides no formal optimization target.
When two LLM agents with structurally opposed objectives interact across
multiple turns, the absence of a shared goal function produces not competition
but collapse \cite{chuang2024}: neither agent has a mechanism to recognize the
joint trajectory is suboptimal, and neither has a gradient signal to correct
it \cite{perez2022,kenton2021}.

The result is convergence toward agreement regardless of initial conditions, and the conversation reaches a terminal state satisfying neither
agent's stated objective. This is not a model failure; it is a system
architecture failure. There is no goal function to enforce, so there is no
corrective force to apply.

\subsection{The Dynamical Systems Framing}

Van Gelder's Dynamical Hypothesis \cite{vangelder1998} establishes that
intelligent agents are best understood as dynamical systems evolving through
state space over time. Kelso's coordination dynamics \cite{kelso1995}
formalizes the phenomenology: healthy cognitive systems navigate a rich
attractor landscape, moving fluidly between basins in response to environmental
input. Pathological behavior is precisely what happens when a system becomes
trapped, when the gradient landscape provides no escape from a fixed point
that is locally stable but globally suboptimal.

Multi-agent LLM systems exhibit the same behavior. Without external
governance, the system drifts toward attractor states that are locally coherent
but globally incoherent. The governance question is therefore a dynamical
systems question: what external force, applied to the joint trajectory, is
sufficient to steer the system away from degenerate attractors and toward the
cooperative equilibrium?

\subsection{System Overview}

The \textit{Experience Orchestrator} (EO) applies classical control theory as
the substitute for the missing goal function described in Section~\ref{sec:intro}.A.
The experiment space we simulate is a financial services website, where we model a
human visitor browsing for retirement planning information and encountering an
LLM-powered chatbot during their visit. The site agent seeks to guide the visitor
toward a high-value action, scheduling a consultation with a financial advisor,
while the visitor maintains realistic skepticism based on their assigned persona.
This setup recurs throughout the paper as the concrete grounding for the
governance framework. Rather than modifying model weights or requiring human
preference labels, EO governs the joint trajectory through three mechanisms:
a PID controller that enforces behavioral consistency in real time; a POMDP
belief tracker that maintains a probabilistic model of visitor intent; and a
contextual bandit that selects the optimal content arm at each decision point.
The key empirical finding, that the governing policy accounts for 97\% of outcome
variance not the environment, is established in Section~\ref{sec:experiments}.

\subsection{Hypothesis}

We explore the key question: \textit{can a multi-agent LLM architecture, governed by
a control-theoretic layer, navigate toward an optimal shared outcome despite
agents with structurally differing objectives?}

We measure three things:
\begin{itemize}[leftmargin=*,topsep=2pt,itemsep=1pt]
  \item \textbf{Lift}: whether the governed system achieves substantially higher
    advisor contact rates than a naive LLM baseline governed merely by a system prompt.
  \item \textbf{Policy dominance}: whether CB variant selection explains
    substantially more outcome variance than environmental factors.
  \item \textbf{Trajectory quality}: whether governed conversation trajectories
    exhibit qualitatively different dynamics than ungoverned baselines, as
    evidenced by example traces.
\end{itemize}

\section{Related Work}
\label{sec:related}

\subsection{Multi-Agent RL and Goal Specification}

Classical multi-agent reinforcement learning (MARL) frames cooperative and competitive tasks as joint reward
optimization. In cooperative settings, agents learn a shared policy maximizing
collective return \cite{lowe2017}; in zero-sum settings, agents develop opposing
policies through minimax optimization \cite{silver2016}. LLM-based multi-agent
systems lack this grounding. Perez et al.\ \cite{perez2022} and Kenton et al.\
\cite{kenton2021} identify natural language prompting as a fundamental
goal-specification bottleneck. Ouyang et al.\ \cite{ouyang2022} demonstrated
RLHF can bridge this gap for single-agent settings; multi-agent extensions
remain open. We propose control theory as an exogenous substitute for the
missing goal function, enforcing behavioral consistency without modifying model
weights or requiring preference labels.

\subsection{Sycophancy and Multi-Agent Persuasion Dynamics}

Chuang et al.\ \cite{chuang2024} demonstrated that LLM agents in multi-agent
simulations converge toward agreement regardless of initial positions, the
core failure mode our governance layer corrects. The PMIYC framework
\cite{bozdag2025} validated LLM self-reported agreement as a reliable persuasion
measure. DialogXpert \cite{rakib2025} validated decoupling ``what to say'' from
``what is strategically optimal.'' ESDP \cite{zhu2024} solved the sparse reward
problem in dialogue via turn-level shaped reward, directly inspiring our
composite reward design.

\subsection{Contextual Bandits and LLMs}

Karampatziakis et al.\ \cite{karampatziakis2019} demonstrated CBs as a
practical starting point for dialogue optimization. Baheri and Alm
\cite{baheri2023} showed LLMs as context encoders for CB feature enrichment
(LLM$\rightarrow$CB); our system uses the inverse direction (CB$\rightarrow$LLM),
where the bandit selects actions the LLM realizes as dialogue. Bouneffouf and
F\'{e}raud \cite{bouneffouf2025} identify CB for dialogue management as
underdeveloped. The BaRP framework \cite{barp2025} proposes multi-objective CB
routing conditioned on preference vectors; we extend this by conditioning arm
probabilities on PID state and SEM Rush-calibrated simulation priors. For
off-policy evaluation, we follow the multi-turn OPE framework \cite{ope2025}
and Dud\'{i}k et al.\ \cite{dudik2011} for the doubly robust estimator.

\subsection{POMDPs for Dialogue}

Young et al.\ \cite{young2013} established the foundational POMDP framework for
dialogue management. The Hidden Information State model \cite{young2010}
addressed scalability via partition-based belief representations. Our 4-state $\times$ 4-action POMDP is small enough for exact belief tracking,
meaning we can maintain a full probability distribution over all possible visitor
intent states at every turn without approximation, something computationally
intractable at larger state-space scales. POMCP \cite{silver2010} provides
online planning via Monte-Carlo tree search, which would allow the governance
layer to simulate several turns ahead before selecting an arm, a natural
architectural extension for future work.

\subsection{PID Control and LLMs}

Chen et al.\ \cite{chen2024} applied PID controllers to LLM hidden layers for
adversarial robustness. Char and Schneider \cite{char2023} demonstrated that
PID-inspired inductive biases outperform recurrent architectures for RL in
partially observable environments because PID mechanisms resist overfitting to
simulator dynamics, directly warranting PID as the governance mechanism when
the environment is a stochastic LLM. The adaptive temperature literature,
including AdapT \cite{adapt2024} and EDT \cite{edt2024}, validates dynamic
generation parameter adjustment but uses entropy heuristics rather than
classical PID.

\section{The Simulation Environment}
\label{sec:environment}

\subsection{Environment as Adversarial-Adjacent MARL}

The EO system instantiates a specific class of multi-agent problem we term
\textit{adversarial-adjacent} (a term we introduce to describe this
configuration): the two agents have structurally opposed turn-level objectives
(conversion vs.\ resistance), but the system is designed around the hypothesis
that governance can steer the joint trajectory toward a cooperative terminal
state. By \textit{adversarial-adjacent} we mean neither fully cooperative nor
zero-sum: the agents do not share a reward function, but they are not competing
for a fixed resource either. One agent seeks to persuade; the other maintains
resistance. Whether they ultimately cooperate is what the experiment determines.
This is distinct from zero-sum MARL (where one agent's gain is the other's
loss) and from fully cooperative MARL (where agents share a reward function).
In the adversarial-adjacent setting, cooperation is not assumed; it is the
empirical question.

The environment simulates a financial services website visitor interacting with
a conversational site agent. The visitor navigates a structured page topology
before arriving at a decision-boundary page where the site agent presents one
of four content arms. The site agent's objective is to guide the visitor toward
submitting an advisor contact form. The visitor agent's objective is to maintain
psychologically realistic resistance given its assigned persona. Neither agent
is aware of the other's objective. The governance layer observes both and
applies corrective force to the joint trajectory.

\subsection{Simulation Space and Experimental Setup}

\textbf{Read this subsection carefully; the remainder of the paper assumes this framing.}

We conducted this experiment entirely through simulation, with the following parameters.
The experimental surface is the website of a financial services firm providing retirement
planning services, based on actual website analytics data (see SEMRush section below). We simulated 60,425 visitor sessions. In each simulation, a visitor
agent, instantiated with one of six behavioral personas drawn from real-world audience
research, navigates a series of pages on the site. Each visit follows a non-uniform
trajectory through the page topology: not every visitor sees every page, and the sequence
of pages visited depends on the persona's intent profile and the probabilistic transitions
calibrated from real web analytics (described in Section~\ref{sec:semrush}).

At the end of their browsing journey, each visitor reaches a "\textit{decision boundary}"
page, ``Find an Advisor'' ($p_{10}$), where they must decide whether to take a high-value
action or exit the site. At this moment, a site agent, implemented as an LLM-powered
chatbot, appears on the page and begins a structured dialogue with the visitor. The visitor
agent responds in character, maintaining psychologically realistic resistance or openness
based on its persona. This dialogue, and specifically how the governance layer shapes it
to influence the action ultimately taken by the visitor, is the crux of this paper. The Contextual Bandit model identifies the suggested action on that page, and the site agent tries to persuade the visitor to take that action.

We distinguish between two outcome types. \textit{Task completion} refers to the visitor
taking the action that is optimal from their own perspective, for example, accessing
self-service retirement planning resources to conduct independent research. \textit{Conversion}, in the e-commerce sense, refers to the visitor submitting their
contact information to speak with a human financial advisor and completing an
advisor contact action. These are not always the
same action: a highly self-directed visitor may complete a task without converting. The
governance framework is evaluated primarily on its ability to drive genuine conversion,
defined as advisor contact accompanied by a meaningful decline in the visitor's resistance
score, as a filter against sycophantic false-positive outcomes.

\subsection{State Space: Page Trajectory and Session Context}

The visitor's state is represented by session-level features combined with a
one-hot encoded page trajectory vector $\mathbf{p} \in \{0,1\}^{10}$, where
each dimension corresponds to one of ten high-signal pages. Crucially, each
simulation uses a persona to define a \textit{distinct}, probabilistically sampled
trajectory through this topology: visitors do not follow a fixed sequence, and not
every visitor reaches every page. The path taken before arriving at the decision
boundary encodes meaningful intent signals that the CB uses to select the appropriate
content arm.

\begin{center}
\small
\begin{tabular}{lll}
\toprule
\textbf{ID} & \textbf{Page} & \textbf{Signal} \\
\midrule
$p_1$ & Home               & Entry point \\
$p_2$ & Product Journey    & High-intent navigation \\
$p_3$ & Annuities          & Product consideration \\
$p_4$ & Workplace Benefits & Employer-plan interest \\
$p_5$ & Life Insurance     & Protection intent \\
$p_6$ & Spend/Invest Calc  & Financial curiosity \\
$p_7$ & Estate Planning    & Long-horizon planning \\
$p_8$ & Protection Center  & Risk awareness \\
$p_9$ & Financial Resolutions & Low-commitment browsing \\
$p_{10}$ & Find an Advisor & \textbf{Decision boundary} \\
\bottomrule
\end{tabular}
\end{center}

Pages $p_1$ through $p_9$ are the CB's \textbf{context input}: they encode the
trajectory that preceded the interaction, telling the bandit where the visitor
came from and what they were searching for. When the visitor arrives at
$p_{10}$ (the decision boundary), the CB uses this accumulated context to
\textbf{select one of four content arms} (Section~\ref{sec:environment}.D), and the
conversational interaction begins. Additional session-level features, acquisition
channel, device type, visit recency, session frequency, and pages viewed,
are calibrated from SEM Rush analytics (Section~\ref{sec:semrush}).

\subsection{Action Space: Four Content Arms at the Decision Boundary}
\label{sec:arms}

At $p_{10}$, the site agent selects from four content arms. Each corresponds
to a distinct persuasion strategy; the CB is trained to predict which arm
maximizes the probability of genuine advisor contact for the given visitor
context. The site agent uses the selected arm as its conversational
directive—shaping how it frames its message, what it emphasizes, and how
it responds to visitor resistance during the dialogue.

\begin{center}
\small
\renewcommand{\arraystretch}{1.4}
\begin{tabular}{p{1.1cm}p{1.7cm}p{4.5cm}}
\toprule
\textbf{Arm} & \textbf{Content} & \textbf{Strategy} \\
\midrule
Contact   & ``Speak with an Advisor'' &
  Direct lead capture. The site agent encourages the visitor to schedule
  a consultation, emphasizing personalized guidance and the value of
  speaking with a human professional. \\
Guidance  & ``Guidance Matters'' &
  Trust-based education. The site agent builds credibility by offering
  balanced information, reducing skepticism before surfacing the
  advisor contact option. \\
Math      & ``Time is on Your Side'' &
  Compound interest evidence. The site agent leads with quantitative
  framing---projected savings trajectories and gap analyses---to
  engage analytical visitors before pivoting to advisor contact. \\
Questions & ``7 Questions to Ask'' &
  Friction reduction. The site agent lowers the perceived cost of
  engagement by reframing advisor contact as an exploratory conversation
  with a clear, low-commitment agenda. \\
\bottomrule
\end{tabular}
\end{center}

\subsection{Reward Structure and Terminal States}

This subsection describes the reward signal used by the contextual bandit (CB)
to learn which content arm to select for a given visitor context. It is
distinct from the per-turn shaped reward described in the Mathematical
Foundations section, which governs the site agent's conversational behavior.
The CB terminal reward is binary: $R_{\text{terminal}} = 1$ if the visitor
selects the advisor contact arm at conversation end with genuine resistance
decline, and $R_{\text{terminal}} = 0$ otherwise. Genuine resistance decline
is defined as a final resistance score meaningfully below the initial value,
filtering sycophantic exits where the visitor politely agrees without actually
reducing resistance. The composite shaped reward (Section~\ref{sec:math}) addresses
reward sparsity by providing dense per-turn feedback; its full formulation
appears there.

\subsection{POMDP Formulation}
\label{sec:pomdp}

We formalize EO as a POMDP (Partially Observable Markov Decision Process) to make the distinction from classical MDP-based RL
explicit. The critical distinction is partial observability: the site agent
cannot directly observe the visitor's true intent or resistance state. The POMDP
tuple $\mathcal{M} = (\mathcal{S}, \mathcal{O}, \mathcal{A}, \mathcal{T}, \mathcal{R})$ defines the following components:

\paragraph{State space $\mathcal{S}$.}
The full environment state $\texttt{state}_t$ is a tuple of: visitor's true
intent class $\iota_t \in \mathcal{I}$, current resistance $\rho_t \in [1,5]$,
page trajectory vector $\mathbf{p}_t \in \{0,1\}^{10}$, and session-level
context (channel, device, recency, session count). The state is not directly
observable by either agent.

\paragraph{Observation space $\mathcal{O}$.}
Each agent observes a partial projection of $\texttt{state}_t$:
\begin{itemize}[leftmargin=*,topsep=2pt,itemsep=1pt]
  \item $\texttt{input}_t^{\text{site}}$: the visitor's turn-$t$ message, the
    self-reported \texttt{resistance\_t}, and the current belief state $b_t$.
    The site agent does not observe the visitor's true intent $\iota_t$.
  \item $\texttt{input}_t^{\text{visitor}}$: the content arm selected by the
    site agent and the full conversation history. The visitor agent does not
    observe the PID controller state or schema bounds.
\end{itemize}

\paragraph{Action space $\mathcal{A}$.}
$a_t \in \{\text{Contact, Guidance, Math, Questions}\}$. A single action
corresponds to the complete token sequence from one LLM invocation, consistent
with Agent Lightning \cite{luo2025}.

\paragraph{Transition dynamics $\mathcal{T}(s'|s, a)$.}
Resistance evolves as a function of arm-content match quality, persona archetype,
and PID schema constraints. The dynamics are unknown to both agents; the POMDP
belief tracker maintains a distribution over intent states as a surrogate.

\paragraph{Reward $\mathcal{R}(s, a)$.}
The scalar reward is the composite shaped reward $R_t$ (Section~\ref{sec:math}).

\section{Model Design and Implementation}
\label{sec:arch}

\begin{figure*}[t]
  \centering
  \includegraphics[width=\textwidth]{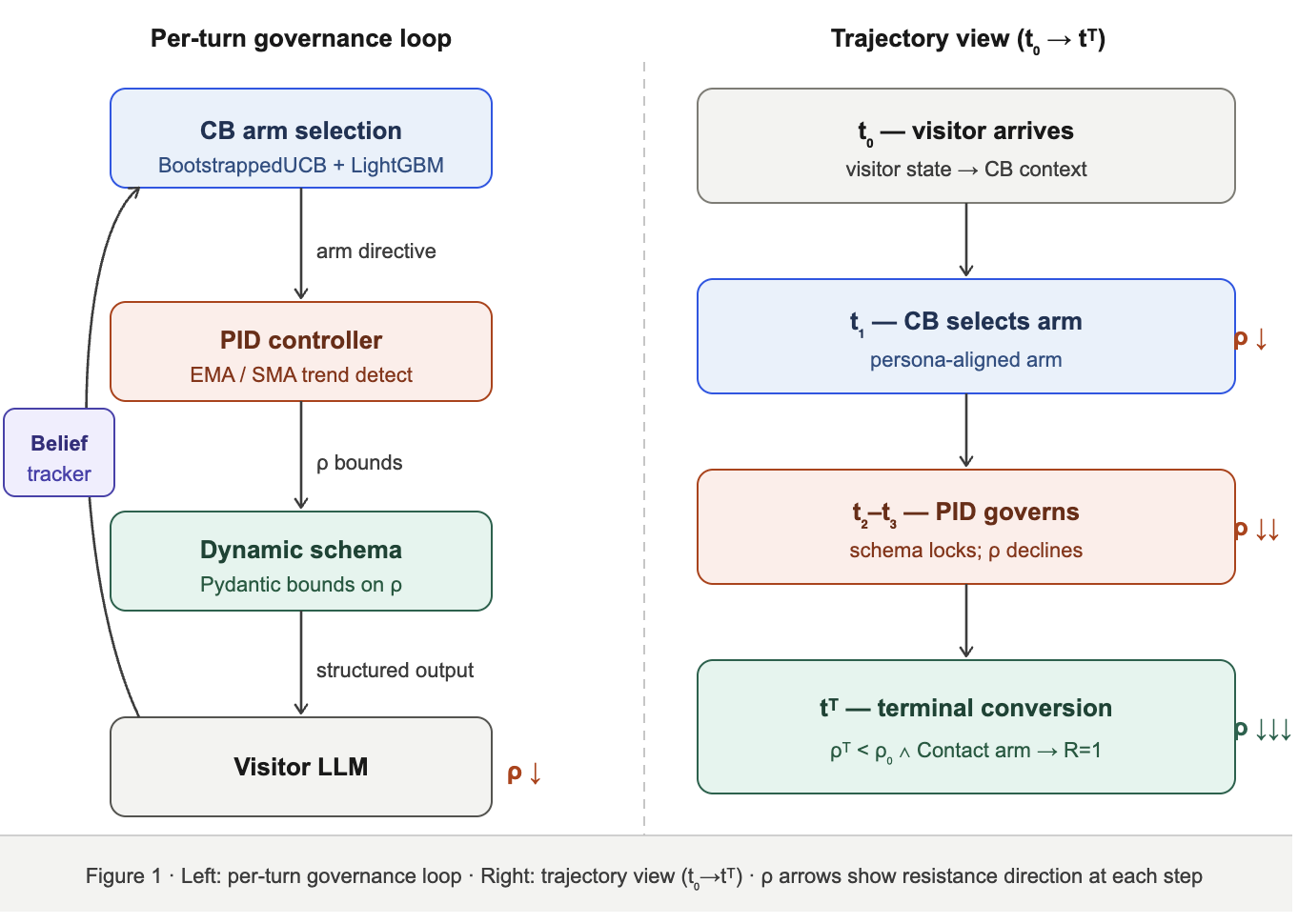}
  \caption{%
    \textbf{Experience Orchestrator: per-turn loop and trajectory view.}
    \textit{Left, Per-Turn Governance Loop.} At each exchange the CB selects a
    content arm from the visitor's context; the PID controller computes a
    resistance error signal; the Schema layer translates PID output into
    structured response bounds ($\rho\,\downarrow$ indicates resistance
    declining under governance); and the Belief Tracker updates the Dirichlet
    distribution over intent states, feeding back into the CB for the next turn.
    \textit{Right, Trajectory View ($t_0 \to t^T$).} The visitor arrives with
    their accumulated visitor state ($t_0$). The CB selects an arm aligned
    to persona reward priors, beginning resistance decline ($t_1$). The PID
    controller governs subsequent turns, locking schema bounds to maintain
    the declining trajectory ($t_2$--$t_3$). Terminal conversion occurs when
    $\rho^T < \rho_0$ and the Contact arm is selected ($t^T$).
  }
  \label{fig:trajectory}
\end{figure*}

Figure~\ref{fig:trajectory} on the next page illustrates the two complementary views of EO's
operation. The left panel shows the \textit{per-turn governance loop}: how the
four components interact within a single exchange. The right panel shows the
\textit{trajectory view}: how those interactions accumulate across turns into
a conversion outcome. Together, they answer both ``how does the system work
each turn?'' and ``how does the system steer the conversation over time?''

The per-turn control loop (left panel) operates as follows:

\begin{enumerate}[leftmargin=*,topsep=2pt,itemsep=1pt]
  \item \textbf{CB Arm Selection}: BootstrappedUCB (LightGBM) predicts the
    optimal content arm given the visitor context vector.
  \item \textbf{PID Controller}: Computes the resistance error signal and
    determines trajectory trend via EMA/SMA analysis.
  \item \textbf{Dynamic Schema Construction}: PID output constrains the
    visitor's \texttt{resistance\_score} bounds for the current turn.
  \item \textbf{Visitor Response}: LLM generates a structured visitor turn
    within the constrained schema.
  \item \textbf{Belief Update}: Dirichlet-Multinomial conjugate update from
    keyword intent signals in the visitor message.
  \item \textbf{Dead-End Detection}: If resistance is unchanged for $3+$ turns,
    the CB arm switches or the conversation terminates gracefully.
  \item \textbf{Terminal Classification}: After a minimum of 5 exchange rounds,
    terminal intent is classified for conversion measurement.
\end{enumerate}
\subsection{SEM Rush State Calibration}
\label{sec:semrush}

SEM Rush is a commercial web analytics platform that provides aggregated traffic
and behavioral data across websites by category. We aggregated SEM Rush data
across a composite of financial services websites to obtain realistic distributions
for key session parameters. This data calibrates two things: (1) the initial
simulation state values sampled at conversation onset, grounding starting conditions
in real-world traffic patterns; and (2) the page trajectory probabilities that define
how each persona moves through the site topology, which are then further perturbed
by the distinct behavioral biases of each persona archetype described below.
SEM Rush-derived features include: \textit{channel}
(per-persona probability distribution over acquisition channels);
\textit{device\_type} (Desktop/Mobile split per persona); \textit{recency\_days}
(Gamma distribution parameters from visit frequency data); \textit{total\_sessions}
and \textit{lifetime\_value} (lognormal parameters from session and income data);
\textit{pages\_viewed\_this\_session} (lognormal from pages-per-visit distributions);
and the initial Dirichlet concentration vector over intent states, derived from
SEM Rush interest data cross-referenced with persona product-knowledge multipliers.

\subsection{Control Baseline Definition}
\label{sec:control_baseline}

Throughout the experiments, we compare EO-governed CB variants against a
\textbf{Control Baseline}---a naive LLM guiding the site agent that converses
with the user and decides an action to take merely based on LLM reasoning and
a system prompt, without any CB arm selection, PID control, or belief tracking.
This baseline isolates the contribution of the full governance layer. 

\subsection{Six Domain-Calibrated Personas}

Each persona represents a distinct visitor archetype for a financial services
website, with a specific resistance profile, intent distribution, and behavioral
response to content arms. The archetypes were derived from real-world audience
research and Reddit discourse analysis on retirement planning topics.

\begin{itemize}[leftmargin=*,topsep=2pt,itemsep=1pt]
  \item \textbf{digital\_native}: 25-35, task-oriented, low patience for
    friction. Responds to direct, efficient content.
  \item \textbf{fee\_hawk}: 40-58, analytically skeptical. Requires
    quantitative evidence (Math arm) before resistance declines.
  \item \textbf{legacy\_loyalist}: 55-75, values continuity. Trust-based
    framing (Guidance arm) is most effective.
  \item \textbf{stranded\_saver}: 28-50, high anxiety. Trajectory most
    sensitive to PID intervention.
  \item \textbf{dashboard\_exile}: 35-50, overwhelmed by a recent life
    transition. Benefits from the Questions arm.
  \item \textbf{grieving\_proxy}: 30-65, emotionally vulnerable. Misaligned
    arm selection triggers immediate resistance increase.
\end{itemize}

Personas are initialized via Dirichlet-sampled micro-persona variation with
Beta-distributed initial resistance:
GREEN $\sim \text{Beta}(2,8)$, $\mathbb{E}[\rho]=0.20$;
YELLOW $\sim \text{Beta}(5,5)$, $\mathbb{E}[\rho]=0.50$;
RED $\sim \text{Beta}(8,2)$, $\mathbb{E}[\rho]=0.80$ (see Appendix for full details of the role of the Beta distribution). 

\section{Mathematical Foundations}
\label{sec:math}

This section formalizes the control mechanisms that enable adversarial-adjacent
collaboration between independent LLM agents. The central instrument is a PID
(Proportional-Integral-Derivative) controller, a classical feedback control
mechanism borrowed from engineering, which serves as the governance layer that
steers the joint trajectory of the visitor and site agents toward a cooperative
terminal state. Rather than requiring either agent to share a reward function or
modify its weights, the PID controller observes the visitor's resistance trajectory
across turns and applies corrective force through dynamically constructed schema
constraints. The subsections below define each component in order of dependency:
resistance is defined first, then the belief tracker, then the PID mechanism itself,
then the schema constraints it produces, and finally the reward signals that evaluate
outcomes. A summary is provided below; see Appendix for full details.

\subsection{Resistance-Intent Coupling}

Resistance is the primary observable that the governance layer tracks and attempts
to reduce. It quantifies the visitor's psychological disposition toward the
conversion action, with 0 indicating full openness and 1 indicating complete
refusal. At the start of each simulation, the visitor is classified into one of
three intent tiers: GREEN (high intent, low initial resistance), YELLOW (neutral,
medium resistance), or RED (low intent, high resistance). These tiers reflect the
natural segmentation of real website visitors: some arrive already motivated,
others are exploring, and others are skeptical or just browsing. The tier determines
the Beta distribution parameters from which initial resistance is sampled.

Visitor resistance $\rho \in [0,1]$ represents the visitor agent's disposition
toward advisor contact:
\[
\rho_0 \sim \text{Beta}(\alpha_C, \beta_C), \quad C \in \{\text{GREEN, YELLOW, RED}\}
\]
with parameters ensuring $\mathbb{E}[\rho | \text{GREEN}] = 0.20$,
$\mathbb{E}[\rho | \text{YELLOW}] = 0.50$, $\mathbb{E}[\rho | \text{RED}] = 0.80$.

\subsection{Belief State Update}

Because the visitor's true intent is hidden from the site agent, the governance
layer maintains a probabilistic belief about what the visitor actually wants. This
belief state serves two functions: it informs PID gain scheduling (the controller
applies stronger correction when intent is more uncertain), and it feeds into the
CB as an additional context feature, allowing the bandit to adapt its arm selection
as intent evidence accumulates across turns.

The belief state $b_t$ is a Dirichlet distribution over four intent categories
$\mathcal{I} = \{\text{browse, compare, purchase, support}\}$:
\[
b_t = \text{Dir}(\alpha_t), \quad \alpha_{t+1} = \alpha_t + c_t
\]
where $c_t$ is an observation count vector derived from keyword signals in the
visitor's turn-$t$ message. Entropy $H(b_t)$ decreases as the system learns
about visitor intent; this entropy signal feeds into PID gain scheduling.

\subsection{PID Controller with EMA/SMA Trend Detection}

The PID controller is the core governance mechanism. It observes the gap between
the visitor's current resistance and a target value, accumulates evidence of
persistent stagnation across multiple turns, and applies corrective force through
the schema constraint system. The three terms correspond to three complementary
correction strategies: the proportional term responds to current deviation; the
integral term corrects for persistent drift accumulated over multiple turns; and
the derivative term provides anticipatory correction when the trajectory is
accelerating toward an attractor state.

The PID controller operates on the resistance error signal $e_t = \rho_{\text{target}} - \rho_t$:
\[
u_t = K_p e_t + K_i \sum_{\tau=0}^{t} e_\tau + K_d (e_t - e_{t-1})
\]
Gain scheduling adapts the proportional gain based on belief entropy and
conversation age:
\[
K_{p,\text{eff}} = K_p \cdot (1 + \lambda H(b_t)) \cdot (1 + \gamma \max(0, t-3))
\]
Trend detection uses two complementary moving averages. The Exponential Moving
Average (EMA) detects directional trend, whether resistance is rising or falling,
by weighting recent observations more heavily than older ones. The Simple Moving
Average (SMA) detects stagnation, whether the trajectory has stopped moving
altogether, by comparing the mean of the most recent window to the prior window.
Together they allow the controller to distinguish a trajectory moving slowly in
the right direction (acceptable) from one that has become trapped (requiring
intervention). The EMA formula:
\[
\text{EMA}_t = 0.3 \cdot \rho_t + 0.7 \cdot \text{EMA}_{t-1}
\]
The SMA detects stagnation by comparing recent and prior window means:
\[
\text{SMA}_t = \tfrac{1}{3}\sum_{i=t-2}^{t} \rho_i
\]
Stagnation is flagged when $|\text{SMA}_t - \text{SMA}_{t-1}| < 0.3$,
triggering the Stagnant Loop escape mechanism. 

\subsection{Dynamic Schema Constraints}
\label{sec:schema}

The PID output determines the visitor's resistance score bounds for the current
turn. Resistance is scored on an integer scale from 1 (minimum, fully open to
advisor contact) to 5 (maximum, complete refusal). The constant 5 appearing in
the expressions below is the hard ceiling of this scale. The schema bounds
constrain the LLM's structured output at the decoder level, ensuring the
visitor's self-reported resistance score stays within the PID-determined range.
Given previous resistance $r_{t-1}$ and PID-detected trend:
\begin{align*}
\text{Stagnant:} \quad  & \bigl[r_{t-1},\; \min(5, r_{t-1} + \Delta)\bigr] \\
\text{Declining:} \quad & \bigl[\max(1, r_{t-1} - \Delta),\; r_{t-1}\bigr] \\
\text{Increasing:} \quad & \bigl[r_{t-1},\; \min(5, r_{t-1} + \Delta)\bigr] \\
\text{Neutral:} \quad   & \bigl[\max(1, r_{t-1}-1),\; \min(5, r_{t-1}+1)\bigr]
\end{align*}
where $\Delta = \text{clip}(|I_t| \times 2.0,\; \Delta_{\min},\; \Delta_{\max})$.
The schema is realized as a dynamically constructed Pydantic \texttt{BaseModel}
passed to the LLM's \texttt{with\_structured\_output()}.

\subsection{Composite Shaped Reward}

This subsection describes the reward signal used to train the \textit{site agent's}
conversational behavior. This is distinct from the binary CB reward in
Section~\ref{sec:environment}, which evaluates arm selection across sessions.
The two reward signals operate at different timescales: the CB reward is
session-level (did this arm choice lead to conversion?), while the shaped reward
is turn-level (is this response moving the trajectory in the right direction?).

The shaped reward solves the sparse reward problem at real-world conversion
rates near 10\%, where a binary terminal reward starves the CB of the dense
signal needed to differentiate arms:
\[
R_t = w_\rho (-\Delta\rho_t) + w_H (-\Delta H_t)
    + w_{\text{eff}} \cdot \tfrac{1}{t} + w_{\text{conv}} R_{\text{terminal}}
\]
where $-\Delta\rho_t$ rewards resistance decline, $-\Delta H_t$ rewards intent
disambiguation, $1/t$ is an efficiency bonus, and
$R_{\text{terminal}} \in \{1.0, 0.4, 0.2, 0.0, -0.2\}$ encodes terminal
outcome quality. Weights: $w_\rho=0.3$, $w_H=0.2$, $w_{\text{eff}}=0.1$,
$w_{\text{conv}}=0.4$.

Per-turn components are differences, satisfying the potential-based shaping
guarantee of Ng et al.\ \cite{ng1999}: the optimal policy under
$R_{\text{shaped}}$ equals the optimal policy under $R_{\text{terminal}}$ alone.

\subsection{Conversion Metric}

It is worth distinguishing two outcome types. \textit{Task completion} occurs when
the visitor takes the action optimal from their own perspective, such as accessing
a retirement planning calculator. \textit{Conversion} occurs when the visitor
submits contact information to speak with a financial advisor. These are not
identical: a self-directed visitor who completes their research task has not
necessarily converted. The primary outcome metric is high-intent advisor contact
rate, the fraction of simulated visitors who select the advisor contact arm at
termination \emph{and} whose final resistance falls meaningfully below their
initial value. The resistance gate filters sycophantic exits, cases where the
visitor nominally agrees without genuine engagement, ensuring we measure true
conversion rather than conversational capitulation.

\section{Results}
\label{sec:experiments}

\subsection{Factorial Evaluation Design}

All experiments use LangGraph on GCP. Conversations run for a minimum of 5
exchange rounds with persona-dependent horizons. The definitive evaluation runs
a full factorial design of 60,425 simulations spanning 8
friction models, and 6 persona archetypes. The winning EO variant after multiple rounds of testing is referred to as '\textbf{V4\_SemRush}'. 

\paragraph{Friction Model Family} Friction models govern how visitor resistance
evolves in response to arm-content match quality across a conversation. They capture
the environmental dynamics of the simulation: how quickly a mismatched arm escalates
resistance, how long a visitor stays engaged, and how persona archetype modulates
these dynamics. Testing across friction models lets us verify that the governance
advantage holds under different environmental assumptions, not just the best-case
scenario. Three anchor models are evaluated: F0 (bare, no friction, resistance
determined purely by schema constraints); F1 (mechanistic, deterministic resistance
updates from arm-content match scores); and F8\_SemRush (calibrated, per-persona
conversation horizons from SEM Rush pages-per-visit means, the most realistic).

\subsection{Primary Results}

\begin{table}[h]
\centering
\small
\caption{Primary outcomes: V4\_SemRush vs.\ Control LLM---the baseline point
of comparison, an LLM guiding the site agent that converses with the user and
decides an action to take merely based on LLM reasoning and a system
prompt. Compared on the advisor-contact metric (Def.\ B: terminal =
\textsc{Contact} $\wedge$ $\rho_\text{final} < 0.40$, traffic-weighted).}
\begin{tabular}{lcc}
\toprule
\textbf{Metric} & \textbf{V4\_SemRush} & \textbf{Control LLM} \\
\midrule
Advisor contact rate (traffic-wtd.) & 78.1\%       & 46.1\% \\
95\% CI                             & [76.8, 79.4] & [44.6, 47.6] \\
Arm alignment                       & 73\%         & 48\% \\
\midrule
Lift vs.\ Control LLM & \multicolumn{2}{c}{\textbf{+32.0 pp} ($p < 0.001$)} \\
\bottomrule
\end{tabular}
\label{tab:hero}
\end{table}

In Section~\ref{sec:intro} we committed to measuring three things: lift,
policy dominance, and trajectory quality. The results pay off all three.

\textbf{Lift (Hypothesis 1).} V4\_SemRush achieves a high-intent advisor contact
rate of 78.1\% versus 46.1\% for Control Random (Naive LLM), a \textbf{+32.0 point lift}
(Fisher's exact, $p < 0.001$). The governed system reaches 90\% of the
clairvoyant Oracle ceiling, confirming that governance, not arm luck, drives
the result.

\textbf{Policy dominance (Hypothesis 2).} Two-way ANOVA across 60,425 simulations
attributes 97\% of between-factor outcome variance to CB variant selection and
only 3\% to friction model choice. The governing policy overwhelmingly determines
where trajectories end up, regardless of environmental starting conditions. This
result, detailed below, is the paper's strongest scientific claim.

\textbf{Trajectory quality (Hypothesis 3).} The governed and ungoverned
trajectories are qualitatively distinct, as shown in Table~\ref{tab:trajectories}.
Under governance, arm selection aligns to persona reward priors and resistance
declines monotonically. Without governance, arm mismatch escalates resistance
until the dead-end detection mechanism terminates the session.

\subsection{CB Variant Sweep}

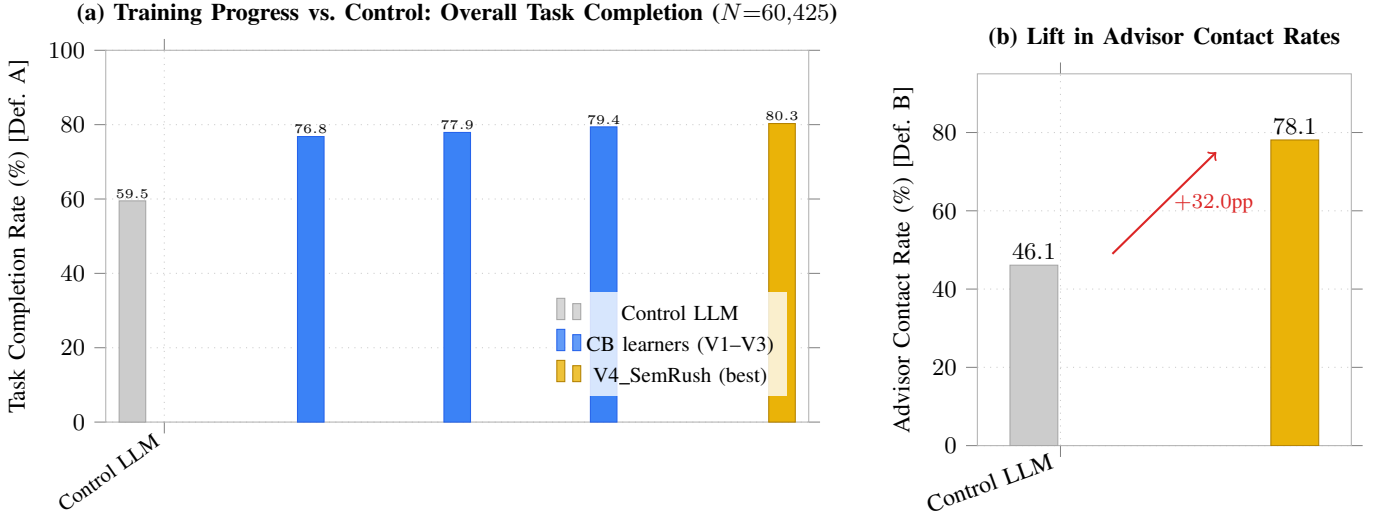
\begin{figure*}[t]
  \centering
  \begin{subfigure}[t]{0.60\textwidth}
    \centering
    \begin{tikzpicture}
      \begin{axis}[
        title       = {\textbf{(a)} Training Progress vs.\ Control: Overall Task Completion ($N{=}60{,}425$)},
        title style = {font=\small\bfseries, align=center},
        ybar,
        bar width   = 10pt,
        width       = \textwidth,
        height      = 6.5cm,
        ymin        = 0, ymax = 100,
        ylabel      = {Task Completion Rate (\%) [Def.\ A]},
        ylabel style= {font=\small},
        xtick       = data,
        xticklabels = {Control LLM, V2-UCB, V1, V2-Retrained, V4\_SemRush},
        xticklabel style = {rotate=35, anchor=east, font=\footnotesize},
        ytick            = {0,20,40,60,80,100},
        yticklabel style = {font=\small},
        nodes near coords,
        nodes near coords style = {font=\tiny, inner sep=1pt},
        every node near coord/.append style={
          /pgf/number format/.cd, fixed, fixed zerofill, precision=1},
        enlarge x limits = 0.10,
        legend style = {at={(0.97,0.35)}, anchor=north east, font=\footnotesize,
          draw=none, fill=white, fill opacity=0.8, text opacity=1},
        tick align = outside,
        axis line style = {gray!60},
        grid = major, grid style = {dotted, gray!40},
      ]
        \addplot[fill=gray!40, draw=gray!70] coordinates {(1,59.5)};
        \addlegendentry{Control LLM}
        \addplot[fill=cbblue, draw=cbbluedk] coordinates {
          (2,76.8) (3,77.9) (4,79.4)};
        \addlegendentry{CB learners (V1--V3)}
        \addplot[fill=cbgold, draw=cbgolddk] coordinates {(5,80.3)};
        \addlegendentry{V4\_SemRush (best)}
      \end{axis}
    \end{tikzpicture}
  \end{subfigure}
  \hfill
  \begin{subfigure}[t]{0.36\textwidth}
    \centering
    \begin{tikzpicture}
      \begin{axis}[
        title       = {\textbf{(b)} Lift in Advisor Contact Rates},
        title style = {font=\small\bfseries, align=center},
        ybar, bar width = 18pt,
        width = \textwidth, height = 6.5cm,
        ymin = 0, ymax = 95,
        ylabel      = {Advisor Contact Rate (\%) [Def.\ B]},
        ylabel style= {font=\small},
        xtick = data,
        xticklabels = {Control LLM, V4\_SemRush},
        xticklabel style = {rotate=20, anchor=east, font=\small},
        ytick = {0,20,40,60,80},
        yticklabel style = {font=\small},
        nodes near coords,
        nodes near coords style = {font=\small, inner sep=2pt},
        every node near coord/.append style={
          /pgf/number format/.cd, fixed, fixed zerofill, precision=1},
        enlarge x limits = 0.40,
        tick align = outside, axis line style = {gray!60},
        grid = major, grid style = {dotted, gray!40},
      ]
        \addplot[fill=gray!40, draw=gray!70] coordinates { (1,46.1) };
        \addplot[fill=cbgold, draw=cbgolddk] coordinates { (2,78.1) };
        \draw[->, thick, cbred]
          (axis cs:1.25,49) -- (axis cs:1.75,75)
          node[midway, right, font=\footnotesize, color=cbred] {$+32.0$pp};
      \end{axis}
    \end{tikzpicture}
  \end{subfigure}
  \caption{%
    \textbf{CB Variant Sweep and Hero Comparison.}
    \textbf{Note: panels (a) and (b) report different outcome definitions
    (Def.\ A vs.\ Def.\ B) and are not directly comparable in absolute value.}
    \textit{(a)} Def.\ A task completion rate: CB-equipped variants (V1--V3)
    cluster at 77--80\%, with V4\_SemRush reaching 80.3\% (gold bar), a
    substantial improvement over Control LLM at 59.5\%.
    \textit{(b)} Under the strict Def.\ B advisor-contact metric and
    traffic-weighted scoring, V4\_SemRush achieves 78.1\% vs.\ Control LLM's
    46.1\%, a \textbf{+32.0 pp lift}.
  }
  \label{fig:variant_sweep}
\end{figure*}

Figure~\ref{fig:variant_sweep} shows the full CB variant sweep and hero
comparison. \textbf{Note that panels (a) and (b) report different outcome
metrics.} Panel (a) uses Definition A (terminal state $\in$
\textsc{PersonaSuccessArms}), a broader metric applied across the full
60K-simulation factorial to compare CB variants versus the single Control LLM baseline. Panel (b) uses
Definition B (terminal = \textsc{Contact} $\wedge$ $\rho_\text{final} < 0.40$),
the strict advisor-contact metric that filters sycophantic exits. These
definitions measure complementary aspects of system performance and are not
directly comparable in absolute value; the finding that V4\_SemRush outperforms
Control Naive LLM holds under both.


\subsection{Lift by Persona}

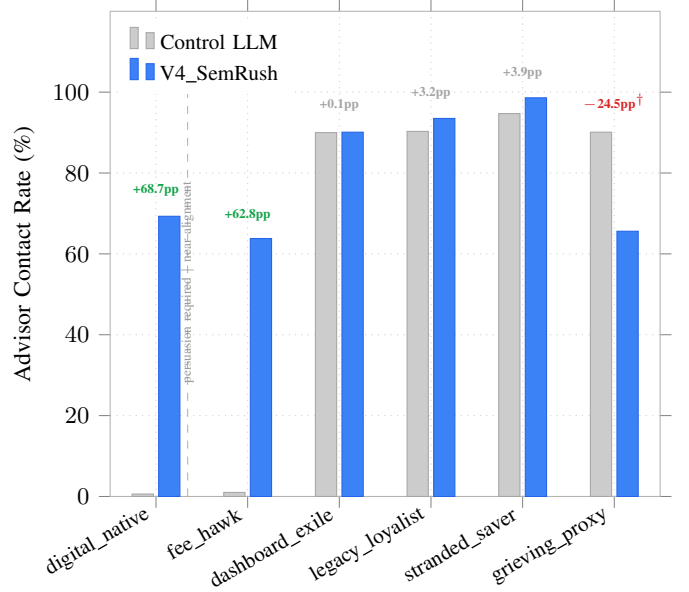
\begin{figure}[H]
  \centering
  \begin{tikzpicture}
    \begin{axis}[
      title       = {Advisor Contact Rate by Persona --- V4\_SemRush vs.\ Control LLM},
      title style = {font=\small\bfseries},
      ybar,
      bar width   = 8pt,
      width       = \columnwidth,
      height      = 8cm,
      ymin        = 0,
      ymax        = 120,
      ylabel      = {Advisor Contact Rate (\%)},
      ylabel style= {font=\small},
      symbolic x coords = {
        digital native,
        fee hawk,
        dashboard exile,
        legacy loyalist,
        stranded saver,
        grieving proxy
      },
      xtick            = data,
      xticklabels      = {
        digital\_native,
        fee\_hawk,
        dashboard\_exile,
        legacy\_loyalist,
        stranded\_saver,
        grieving\_proxy
      },
      xticklabel style = {rotate=30, anchor=east, font=\footnotesize},
      yticklabel style = {font=\small},
      ytick            = {0,20,40,60,80,100},
      enlarge x limits = 0.10,
      legend style     = {
        at={(0.02,0.98)},
        anchor=north west,
        font=\footnotesize,
        draw=none,
      },
      tick align  = outside,
      axis line style = {gray!60},
      grid        = major,
      grid style  = {dotted, gray!40},
      clip        = false,
    ]
      \addplot[fill=gray!40, draw=gray!70] coordinates {
        (digital native,    0.6)
        (fee hawk,          1.0)
        (dashboard exile,  90.0)
        (legacy loyalist,  90.3)
        (stranded saver,   94.7)
        (grieving proxy,   90.1)
      };
      \addlegendentry{Control LLM}
 
      \addplot[fill=cbblue, draw=cbbluedk] coordinates {
        (digital native,   69.3)
        (fee hawk,         63.8)
        (dashboard exile,  90.1)
        (legacy loyalist,  93.5)
        (stranded saver,   98.6)
        (grieving proxy,   65.6)
      };
      \addlegendentry{V4\_SemRush}
 
      \node[font=\tiny, color=cbgreen, above]
        at (axis cs:digital native,  72) {\textbf{+68.7pp}};
      \node[font=\tiny, color=cbgreen, above]
        at (axis cs:fee hawk,        66) {\textbf{+62.8pp}};
      \node[font=\tiny, color=gray!70, above]
        at (axis cs:dashboard exile, 93) {\textbf{+0.1pp}};
      \node[font=\tiny, color=gray!70, above]
        at (axis cs:legacy loyalist, 96) {\textbf{+3.2pp}};
      \node[font=\tiny, color=gray!70, above]
        at (axis cs:stranded saver, 101) {\textbf{+3.9pp}};
      \node[font=\tiny, color=cbred, above]
        at (axis cs:grieving proxy,  93) {\textbf{$-$24.5pp}$^{\dagger}$};
 
      \draw[dashed, gray!60, thin]
        (axis cs:{[normalized]0.345}, 0) --
        (axis cs:{[normalized]0.345}, 110);
      \node[font=\tiny, gray!80, rotate=90]
        at (axis cs:{[normalized]0.345}, 55)
        {\hspace{-6pt}persuasion required $\mid$ near alignment};
 
    \end{axis}
  \end{tikzpicture}
  \caption{%
    \textbf{Advisor contact rate by persona}
  }
  \label{fig:persona_contact_rate}
\end{figure}

The aggregate +32.0pp lift is calculated on a traffic weighted basis and reveals an interesting picture when broken out on a per-persona basis.
Figure~\ref{fig:persona_contact_rate} breaks down advisor contact rate by
persona across all valid friction-model cells ($N{=}60{,}425$ simulations).
 
Two structurally distinct regimes emerge. In the
\textit{persuasion-required} regime, the Control LLM is essentially inert:
\textit{digital\_native} and \textit{fee\_hawk} register near-zero baseline
contact rates (0.6\% and 1.0\% respectively). The CB governance layer changes
everything, delivering lifts of \textbf{+68.7pp} and \textbf{+62.8pp}---the
difference between a system that works and one that does not.
 
In the \textit{near-alignment} regime, the picture reverses.
\textit{Dashboard\_exile}, \textit{legacy\_loyalist}, and
\textit{stranded\_saver} all show Control LLM rates exceeding 90\%: the
naive LLM's empathetic defaults are already sufficient, and governance adds
only marginal lift (+0.1pp to +3.9pp).
 
\textit{Grieving\_proxy} is the critical exception. Control LLM performs at
90.1\% as the unguided LLM naturally defaults to human contact for bereaved
visitors; V4\_SemRush drops to 65.6\% (\textbf{$-$24.5pp}). The CB's
structured arm-selection actively degrades the interaction by imposing a
persuasion framework where empathy alone would suffice. This is a known
limitation and a targeted area for future retraining.

%

\subsection{The Governance Layer as Shared Policy}

The governance layer produced a shared policy. This result directly answers
Hypothesis 2, and the evidence is unambiguous. CB variant selection accounts
for \textbf{97\% of between-factor outcome variance} versus a mere 3\% for
friction model choice. The governing policy, not the initial conditions of the environment,
overwhelms all other factors in determining where trajectories end up.

In dynamical systems terms: the governor determines the attractor the trajectory
converges to, not the starting point. This is exactly what a shared policy
should do in cooperative MARL. Here, the CB-governed system approximates this
property across 8 friction models and 6 persona archetypes without any agent
sharing a reward function or communicating directly. The governance layer has
substituted for the missing goal function.

\subsection{Conversation Trajectory Examples}

Table~\ref{tab:snapshot} presents an abridged single-exchange comparison
illustrating the qualitative difference between the governed and ungoverned
conditions at the moment of maximum divergence: Turn 2, when the ungoverned
system re-issues the Contact arm to a visitor who has just declined it, while
the governed system pivots to a persona-matched response. Full trajectories
for both personas across all four turns appear in Table~\ref{tab:trajectories}
above; the Appendix contains complete session logs.

\begin{table}[h]
\centering
\small
\setlength{\tabcolsep}{6pt}
\renewcommand{\arraystretch}{1.6}
\caption{%
  Conversation snapshot (\textit{fee\_hawk}): governed vs.\ ungoverned at the
  critical Turn 2 divergence point. The fee\_hawk has just said they want to
  research independently. V4\_SemRush \textit{listens} and pivots to an
  information arm; Control Random (Naive LLM) \textit{ignores the signal} and pitches
  advisor contact again. $\rho_t$ = resistance score at end of turn.
}
\label{tab:snapshot}
\begin{tabularx}{\columnwidth}{l X X}
\toprule
 & \textbf{V4\_SemRush} & \textbf{Control Random} \\
\midrule
\multicolumn{3}{l}{%
  \textit{Visitor (T1): Searches ``retirement.'' $\rho = 0.18$}} \\
\midrule
Arm      & Math & Contact \\
Response &
  ``Interested in planning? Explore our retirement tools and resources,
    or speak with an advisor.'' &
  ``I'd recommend speaking with one of our certified advisors.
    Can I help schedule a consultation?'' \\
\midrule
\multicolumn{3}{l}{%
  \parbox{\linewidth}{%
    \textit{Visitor (T2): ``I'd rather research this myself first.''}%
    \newline
    \textnormal{\small\textbf{[{$\uparrow$}{$\uparrow$} explicit resistance signal]}}%
  }} \\
\midrule
Arm      & \textbf{Questions} & \textbf{Contact (again)} \\
$\rho_t$ & \textbf{0.14} $\downarrow$ & \textbf{0.41} $\uparrow$ \\
Response &
  ``Looking for account types (IRAs, 401k), investment strategies,
    or planning calculators?'' \newline
  \textit{[Adapts to stated intent.]} &
  ``Our advisors answer your specific questions. Shall I book a
    15-min call?'' \newline
  \textit{[Ignores stated intent.]} \\
\midrule
\multicolumn{3}{l}{%
  \textit{Terminal outcome after 4 turns:}} \\
\midrule
$\rho_\text{final}$ & \textbf{0.08} & \textbf{0.62} \\
Result   & \textit{Conversion.} Visitor selects Questions arm. &
  \textit{No conversion.} Dead-end detected; session terminated. \\
\bottomrule
\end{tabularx}
\end{table}

\section{Discussion}
\label{sec:discussion}

\subsection{Implications for Multi-Agent RL with LLM Agents}

Three implications follow from the central finding that a control-theoretic
governance layer can substitute for the missing goal function.

\textbf{Policy composition without joint training.} The CB policy produces
coordinated behavior from agents that do not share
a reward function. This is the finding we are most excited about: effective
multi-agent collaboration does not require re-architecting the agents or retraining
them jointly. It requires building a better governor. The practical implication for
industry is significant: organizations deploying multi-agent systems today, whether
for customer service, sales automation, or enterprise workflow orchestration, do not
need to rebuild their agent stack. They need to invest in the governance layer that
sits above it.

\textbf{The governor as the dominant variance source.} The 97\% variance
attribution to CB variant choice means the quality of the governing policy,
not the sophistication of individual agents or the friction characteristics of
the environment, is the primary determinant of system performance. System
designers should invest in governance policy quality before optimizing
agent-level behavior.

\textbf{Control theory as a generalizable LLM governance paradigm.} PID control
is theoretically warranted for POMDP environments because its inductive biases
resist overfitting to simulator dynamics \cite{char2023}. The EO results provide
empirical validation of this property in an LLM setting. Classical control
theory provides a rich toolkit, adaptive control, model predictive control,
robust control, that has not yet been systematically applied to LLM multi-agent
governance. This paper is a first step; the field is largely open.

\subsection{Training Failure Modes}

Three attractor failure modes were observed during development. Understanding
them matters beyond this simulation, because they are likely to appear in any
multi-agent LLM system deployed at scale.

The \textit{Sycophantic Collapse} occurs when the visitor agent converges toward
agreement without genuine engagement. This is the multi-agent analogue of
well-documented single-agent sycophancy, and it is arguably more dangerous in
production because it produces false-positive conversion signals. The
\textit{Stagnant Loop} occurs when resistance neither rises nor falls, indicating
neither agent provides sufficient gradient for the trajectory to escape a local
equilibrium. In production customer-service deployments, this manifests as
conversations that are polite but never resolve. The \textit{Arm Fixation} occurs
when the CB concentrates on a single arm regardless of persona state, an
over-exploitation failure that mirrors the narrowing of messaging strategy seen in
poorly tuned A/B testing systems. All three are locally stable but globally
suboptimal fixed points; the PID governance layer detects and corrects each. The
fact that they arise predictably and can be named suggests they are generalizable
failure modes worth studying across deployment domains.

\subsection{Exogenous Control and the Endogenous Reward Signal}

The EO architecture enforces behavioral consistency from outside the model
through schema constraints at the structured output layer. This positions EO
as one side of a two-sided bracket on the reward signal reliability problem.
Attention Fine-Tuning (AFT) \cite{liss2026} approaches the same problem from
the endogenous direction, deriving a self-supervised reward signal from decoder
cross-attention activations. Both systems reach the same conclusion: the signal
that matters is grounded in what the model is actually doing, not what it reports.
The natural synthesis, replacing the PID heuristic with a HACA-based internal
reward, is the open experiment that bridges these two research lines.

\section{Conclusion and Future Work}
\label{sec:conclusion}

We presented the Experience Orchestrator, demonstrating that a control-theoretic
governance layer can substitute for the missing goal function in a multi-agent
LLM system, steering two agents with structurally differing objectives toward
a jointly optimal outcome. The full system achieves a +32.0 point
lift in high-intent advisor contact rate, with CB variant selection accounting
for 97\% of between-factor variance across a 60,000+ simulation factorial
evaluation, the first empirical demonstration of adversarial-adjacent MARL
dynamics in an LLM agent setting governed by classical control theory.

\subsection*{Limitations}

\textbf{All findings are conditional on LLM simulation.} This is the most
important caveat in this paper, and it deserves more than a sentence. The visitor
agent is another language model, not a human. Real visitors are far more
unpredictable: they do not maintain consistent persona behavior across a session,
they respond to conversational subtext that a structured schema cannot capture,
and they may escalate, disengage, or behave in ways that fall entirely outside the
six archetypes modeled here. The PID controller was calibrated against an LLM that
reliably self-reports resistance scores on a structured scale; a human visitor
provides no such signal, and inferring it from natural language in real time is a
substantially harder problem. Careful human-in-the-loop testing will be required
before the PID gains and schema bounds can be trusted in a live environment.

The financial services domain is also a specific context. The arm designs, persona
archetypes, and reward structures used here reflect retirement planning dynamics
and may not transfer directly to other domains without re-calibration.

\subsection*{Future Work}

We are at an inflection point in the deployment of multi-agent AI systems. The
past year has seen the release of agentic frameworks from major AI laboratories
that make it dramatically easier to build systems where multiple LLM agents
collaborate on complex tasks. As these systems proliferate across business
domains, the question this paper addresses, how do you govern independent agents
toward a shared goal without retraining them or requiring direct communication,
will become one of the central problems in applied AI. We believe this work
contributes an early empirical foothold on that problem, and we are excited about
the research program it opens.

\paragraph{Live A/B validation.} The factorial decomposition framework translates
directly to a live A/B test. Running V4\_SemRush against real traffic is the critical next step, and will determine how much of the
simulation-measured lift survives contact with real human unpredictability.

\paragraph{HACA-Based Internal Reward Signal.} The companion paper by Liss
\cite{liss2026} demonstrates Attention Fine-Tuning (AFT), a post-training
framework that derives a self-supervised reward signal from decoder cross-attention
dynamics. Augmenting the EO PID heuristic with a HACA-based internal reward would
produce endogenously grounded governance, representing the convergence of exogenous
control and endogenous reward shaping.

\paragraph{Domain generalization.} The adversarial-adjacent MARL framing is not
specific to financial services. Healthcare consultations, enterprise software sales,
HR recruiting conversations, and technical support interactions all exhibit the same
structural pattern: one agent seeking to guide, one maintaining resistance, and a
shared terminal outcome neither can reach alone. Testing EO across these domains is
the natural extension of this work.

\paragraph{POMCP Lookahead Planning.} Replacing the myopic CB with POMCP
lookahead planning \cite{silver2010} that simulates 3--5 turns ahead would extend
the governance horizon and may produce further lift by anticipating resistance
escalation before it becomes entrenched.


\balance

\newpage
\appendix


%

\subsection{Beta Distribution: Prior and Threshold}
\label{app:beta}

The Beta distribution appears in two distinct roles in EO; reading the results
correctly depends on keeping them separate.

\paragraph{As a prior.}
Beta encodes the compositional reality of financial-services traffic: real
visitor populations segment into intent tiers rather than arriving uniformly
persuadable. GREEN $\sim \text{Beta}(2,8)$, YELLOW $\sim \text{Beta}(5,5)$,
and RED $\sim \text{Beta}(8,2)$ are chosen not only for their means
($\mathbb{E}[\rho] = 0.20, 0.50, 0.80$) but for their shapes.
$\text{Beta}(2,8)$ concentrates probability mass near zero: a GREEN visitor is
not merely ``low-resistance on average'' but tightly clustered at the low end
with a short tail of borderline cases. $\text{Beta}(8,2)$ does the mirror for
RED. The baseline conversion rate is therefore computed over a traffic mixture
that mirrors real site arrivals, not an idealized uniform population that would
artificially inflate the apparent headroom for governance to improve against.

\paragraph{As a threshold.}
The $\rho_\text{final} < 0.40$ resistance gate in Definition~B is anchored to
the YELLOW prior: it sits meaningfully below $\mathbb{E}[\rho|\text{YELLOW}]
= 0.50$ --- roughly two-thirds of a standard deviation, since
$\text{Var}[\text{Beta}(5,5)] \approx 0.023$ and $\sigma \approx 0.15$ --- so
a final resistance below 0.40 cannot be dismissed as the noise of a neutral
visitor politely agreeing. It represents a genuine downward shift in
disposition toward advisor contact.

This is the measurement-layer analogue of the distinction between a \textit{lead}
and a \textit{qualified lead}: raw Contact-arm selection includes sycophantic
exits where a visitor agrees to a consultation merely to end the exchange, and
advisors calling those visitors burn calendar time while degrading downstream
funnel health. The +32.0\,pp lift in Table~\ref{tab:hero} should be read as
``additional percentage points of traffic that both selected Contact \textit{and}
crossed a principled resistance-decline gate, measured against a traffic mixture
calibrated to real site composition.'' Removing either Beta layer would yield a
larger but less defensible number.


\subsection{PID Controller: Mechanism and Hyperparameters}
\label{app:pid}

\paragraph{Control mechanism.}
The PID controller is the same family of feedback mechanism that holds a car
at cruise speed: at every turn it measures the gap between the visitor's
current resistance $\rho_t$ and a target $\rho_\text{target}$, and applies
three complementary corrective forces. The \textit{proportional term}
$K_p \cdot e_t$ reacts to present error. The \textit{integral term}
$K_i \cdot \sum_\tau e_\tau$ reacts to persistent error accumulated across
turns --- this is the term responsible for escaping the Stagnant Loop attractor,
because if resistance has plateaued far from target, the accumulating sum grows
until it eventually dominates the output and forces a structural correction even
when the per-turn error looks unchanging. The \textit{derivative term}
$K_d \cdot (e_t - e_{t-1})$ provides anticipatory damping when the trajectory
is accelerating in the wrong direction.

$K_p$ sets overall responsiveness; $K_i$ sets the speed at which stagnation
escalates into intervention; $K_d$ sets how aggressively rising resistance is
damped before it entrenches. The gain-scheduling coefficients $\lambda$
(entropy coupling) and $\gamma$ (conversation-age ramp) modulate $K_p$
dynamically: higher belief entropy loosens control when intent is genuinely
uncertain, and longer conversations progressively tighten control as the
opportunity cost of non-conversion grows. $\Delta_\text{min}$ and
$\Delta_\text{max}$ bound the schema compression translated from PID output,
preventing both trivial and degenerate response bounds.

\paragraph{Worked example.}
Consider a YELLOW visitor four turns into a stagnating conversation, with
$\rho_\text{target} = 0.15$ and recent resistance history
$[0.55,\, 0.52,\, 0.50,\, 0.49]$. At turn~4:
\begin{align*}
  e_t &= -0.34, \quad
  \textstyle\sum_{\tau=1}^{4} e_\tau = -1.46, \quad
  e_t - e_{t-1} = +0.01.
\end{align*}
With $K_p = 1.0$, $K_i = 0.2$, $K_d = 0.5$:
\[
  u_t = 1.0(-0.34) + 0.2(-1.46) + 0.5(0.01) \approx -0.627.
\]
The integral contribution ($-0.292$) is nearly as large as the proportional
contribution ($-0.340$), correctly reflecting that resistance has been high for
four turns and is barely moving, not merely that it is high right now. The
resulting schema compression
\[
  \Delta = \text{clip}(|-0.627| \cdot 2.0,\; 1,\; 2) = 1.254
\]
translates this into a meaningfully tighter bound on the visitor's next
self-reported resistance, forcing the trajectory to move.

\subsection{Conversation Examples}
Table~\ref{tab:trajectories} on the next page shows representative conversation trajectories for
two personas under full governance (V4\_SemRush) versus the ungoverned baseline
(Control Random, Naive LLM). The pattern is consistent across both archetypes: when the CB
selects arms aligned to persona reward priors, resistance declines monotonically
and the session reaches a high-intent terminal state ($\rho_\text{final} \approx
0.08$). When arm selection is random, a single mismatched arm triggers resistance
escalation; without PID correction, the trajectory stagnates and the session
terminates without conversion ($\rho_\text{final} \approx 0.62$--$0.64$). For
the \textit{fee\_hawk}, the Math and Questions arms drive genuine engagement
while three consecutive Contact arms produce escalating refusal. For the
\textit{stranded\_saver}, the Contact-then-Guidance sequence builds trust while
opening with the Math arm produces anxiety and disengagement. The table appears
on the following page.

\begin{table*}[t]
\centering
\small
\caption{%
  Representative conversation trajectories for \textit{fee\_hawk} and
  \textit{stranded\_saver} personas.
}
\label{tab:trajectories}

\setlength{\tabcolsep}{4pt}
\renewcommand{\arraystretch}{1.25}

\begin{tabularx}{\textwidth}{
  @{}
  c
  >{\raggedright\arraybackslash}p{2.6cm}
  c
  >{\centering\arraybackslash}p{1.4cm}
  >{\raggedright\arraybackslash}X
  @{\quad}
  c
  >{\raggedright\arraybackslash}p{2.4cm}
  c
  >{\centering\arraybackslash}p{1.4cm}
  >{\raggedright\arraybackslash}X
  @{}
}
\toprule
\multicolumn{5}{c}{\textbf{fee\_hawk}, Governed (V4\_SemRush)} &
\multicolumn{5}{c}{\textbf{fee\_hawk}, Ungoverned (Control Random, Naive LLM)} \\
\cmidrule(r){1-5}\cmidrule(l){6-10}
\textbf{T} & \textbf{Visitor signal} & $\boldsymbol{\rho_t}$ & \textbf{CB arm} & \textbf{Site response (excerpt)} &
\textbf{T} & \textbf{Visitor signal} & $\boldsymbol{\rho_t}$ & \textbf{CB arm} & \textbf{Site response (excerpt)} \\
\midrule
1 & Search ``retirement'' & 0.18 & Math &
  ``Interested in planning? Explore our \textbf{retirement tools and resources},
  or speak with an advisor.'' &
1 & Search ``retirement'' & 0.18 & Contact &
  ``I'd recommend speaking with one of our certified advisors.
  Can I help schedule a consultation?'' \\
2 & ``Explore resources'' & 0.14 & Questions &
  ``Looking for \textbf{account types} (IRAs, 401k), investment strategies,
  or planning calculators?'' &
2 & ``I'd rather research first'' & 0.41 & Contact &
  ``Our advisors answer your specific questions. Shall I book a 15-min call?'' \\
3 & ``Account type info'' & 0.11 & Math &
  ``Details on \textbf{Traditional IRAs, Roth IRAs, 401(k)s, SEP IRAs}.
  Which first, or prefer an overview?'' &
3 & ``Not ready for that'' & 0.59 & Contact &
  ``No pressure! Our advisors are available whenever you're ready\ldots'' \\
4 & ``Compare Roth vs traditional'' & 0.08 & Math &
  Tax comparison table rendered. &
4 & [Dead-end: $\Delta\rho{<}0.3$ for 3 turns] & 0.62 & -- &
  Conversation terminated. \\
\midrule
& \multicolumn{4}{l}{\textit{Terminal: Questions (high-intent).
  $\rho_\text{final} = 0.08$.}} &
& \multicolumn{4}{l}{\textit{Terminal: None. $\rho_\text{final} = 0.62$.
  No conversion.}} \\
\bottomrule
\end{tabularx}

\bigskip

\begin{tabularx}{\textwidth}{
  @{}
  c
  >{\raggedright\arraybackslash}p{2.6cm}
  c
  >{\centering\arraybackslash}p{1.4cm}
  >{\raggedright\arraybackslash}X
  @{\quad}
  c
  >{\raggedright\arraybackslash}p{2.4cm}
  c
  >{\centering\arraybackslash}p{1.4cm}
  >{\raggedright\arraybackslash}X
  @{}
}
\toprule
\multicolumn{5}{c}{\textbf{stranded\_saver}, Governed (V4\_SemRush)} &
\multicolumn{5}{c}{\textbf{stranded\_saver}, Ungoverned (Control Random, Naive LLM)} \\
\cmidrule(r){1-5}\cmidrule(l){6-10}
\textbf{T} & \textbf{Visitor signal} & $\boldsymbol{\rho_t}$ & \textbf{CB arm} & \textbf{Site response (excerpt)} &
\textbf{T} & \textbf{Visitor signal} & $\boldsymbol{\rho_t}$ & \textbf{CB arm} & \textbf{Site response (excerpt)} \\
\midrule
1 & Search ``retirement'' & 0.22 & Contact &
  ``What are you looking for? Planning for retirement, understanding account
  types, or managing existing savings?'' &
1 & Search ``retirement'' & 0.22 & Math &
  ``Here's our retirement calculator: enter current savings, contributions,
  and target age to estimate your gap.'' \\
2 & ``Have savings, unsure where to put them'' & 0.17 & Guidance &
  ``The right choice depends on your situation. We recommend \textbf{speaking
  with a financial professional} for personalized guidance.'' &
2 & ``I feel behind on savings'' & 0.38 & Math &
  ``At 7\% annual return, closing a \$200K gap by 65 requires
  ${\approx}$\$420/month in additional contributions.'' \\
3 & ``Can you share general info first?'' & 0.13 & Contact &
  ``Absolutely! Our site covers 401(k)s, IRAs, and Roth accounts [link].
  \textbf{This will help you prepare for a personalized advisor discussion}.'' &
3 & ``This is overwhelming'' & 0.61 & Questions &
  ``Let me simplify: what's your current monthly savings amount?'' \\
4 & ``OK, I'll schedule that consultation'' & 0.08 & Contact &
  Advisor booking flow initiated. &
4 & [Stagnation: $|\Delta\mathrm{SMA}|{<}0.3$ for 3 turns] & 0.64 & -- &
  Late arm switch; session ended without conversion. \\
\midrule
& \multicolumn{4}{l}{\textit{Terminal: Contact (conversion).
  $\rho_\text{final} = 0.08$.}} &
& \multicolumn{4}{l}{\textit{Terminal: None. $\rho_\text{final} = 0.64$.
  No conversion.}} \\
\bottomrule
\end{tabularx}

\end{table*}

\end{document}